\documentclass[letterpaper, 10 pt, journal]{ieeetran}  

\IEEEoverridecommandlockouts                              

\usepackage{siunitx}
\usepackage{graphics} 
\usepackage{epsfig} 
\usepackage{amsmath} 
\usepackage{amssymb}  
\usepackage{multirow}
\usepackage{algorithm}
\usepackage[]{algpseudocode}

\algnewcommand{\LeftComment}[1]{\Statex \textit{#1}}
\usepackage{nicefrac}
\usepackage{tikz}
\usetikzlibrary{shapes,arrows}
\usetikzlibrary{fit}
\usetikzlibrary{matrix}
\definecolor{colorPink}{rgb}{1,0,1}
\definecolor{colorCyan}{rgb}{0,1,1}

\usepackage{dblfloatfix}

\usepackage[font=small]{subfig}

\title{Multispecies fruit flower detection using a refined semantic segmentation network
}

\author{Philipe A. Dias$^{1}$, Amy Tabb$^{2}$, and Henry Medeiros$^{1}$
\thanks{Manuscript received: February, 24, 2018; Revised May, 17, 2018; Accepted June, 15, 2018.}
\thanks{This paper was recommended for publication by Cyrill Stachniss upon evaluation of the Associate Editor and Reviewers' comments.} 
\thanks{We acknowledge the support of USDA-ARS agreement \#584080-5-020, and of NVIDIA Corporation with the donation of the GPU used for this research.}
\thanks{*Mention of trade names or commercial products in this publication is solely for the purpose of providing specific information and does not imply recommendation or endorsement by the U.S. Department of Agriculture.  USDA is an equal opportunity provider and employer.}
\thanks{$^{1}$P.A. Dias and H. Medeiros are with the Department of Electrical and Computer Engineering,   
        Marquette University, Milwaukee, WI, USA,
        {\tt\footnotesize philipe.ambroziodias@marquette.edu ; henry.medeiros@marquette.edu}}%
\thanks{$^{2}$A. Tabb is with the U.S. Department of Agriculture (USDA), 
        Kearneysville, WV, USA
        {\tt\footnotesize amy.tabb@ars.usda.gov}}%
\thanks{Digital Object Identifier (DOI): 10.1109/LRA.2018.2849498}
\thanks{The citation information for this paper is: P. A. Dias, A. Tabb and H. Medeiros, "Multispecies Fruit Flower Detection Using a Refined Semantic Segmentation Network," in IEEE Robotics and Automation Letters, vol. 3, no. 4, pp. 3003-3010, Oct. 2018.}
}

\begin{document}

\maketitle

\begin{abstract}

In fruit production, critical crop management decisions are guided by bloom intensity, i.e., the number of flowers present in an orchard. Despite its importance, bloom intensity is still typically estimated by means of human visual inspection. Existing automated computer vision systems for flower identification are based on hand-engineered techniques that work only under specific conditions and with limited performance. This work proposes an automated technique for flower identification that is robust to uncontrolled environments and applicable to different flower species. Our method relies on an end-to-end residual convolutional neural network (CNN) that represents the state-of-the-art in semantic segmentation. To enhance its sensitivity to flowers, we fine-tune this network using a single dataset of apple flower images. Since CNNs tend to produce coarse segmentations, we employ a refinement method to better distinguish between individual flower instances. Without any pre-processing or dataset-specific training, experimental results on images of apple, peach and pear flowers, acquired under different conditions demonstrate the robustness and broad applicability of our method.

\end{abstract}


\section{INTRODUCTION}

\IEEEPARstart{B}loom intensity corresponds to the number of flowers present in orchards during the early growing season. Climate and bloom intensity information are crucial to guide the processes of pruning and thinning, which directly impact fruit load, size, coloration, and taste \cite{Forshey1986,Link2000}. Accurate estimates of bloom intensity can also benefit packing houses, since early crop-load estimation greatly contributes to optimizing postharvest handling and storage processes. 

Visual inspection is still the dominant approach for bloom intensity estimation in orchards, a technique which is time-consuming, labor-intensive and prone to errors \cite{Gongal2016}. Since only a limited sample of trees is inspected, the extrapolation to the entire orchard relies heavily on the grower's experience. Moreover, it does not provide information about the spatial variability in the orchard, although the benefits of precision agriculture practices are well known \cite{Zhang2002precision}. 

These limitations added to the short-term nature of flower appearance until petal fall make an automated method highly desirable. Multiple automated computer vision systems have been proposed to solve this problem, but most of these methods rely on hand-engineered features \cite{Kapach2012}, making their overall performance acceptable only under relatively controlled environments (e.g. at night with artificial illumination). Their applicability is in most cases species-specific and highly vulnerable to variations in lightning conditions, occlusions by leaves, stems or other flowers \cite{Gongal2015}.

In the last decade, deep learning approaches based on convolutional neural networks (CNNs) led to substantial improvements in the state-of-the-art of many computer vision tasks \cite{Guo2016cnn}. Recent works have adapted CNN architectures to agricultural applications such as fruit quantification \cite{Chen2017}, classification of crops \cite{Dyrmann2016crops}, and plant identification from leaf vein patterns \cite{Grinblat2016vein}. To the best of our knowledge, our work in \cite{Dias2018} was the first to employ CNNs for flower detection. In that work, we combined superpixel-based region proposals with a classification network to detect apple flowers. Limitations of that approach are intrinsic to the inaccuracies of superpixel segmentation and the network architecture.

In the present work, we provide the following contributions for automated flower segmentation:
\begin{itemize}
    \item A novel technique for flower identification that is i) automated, ii) robust to clutter and changes in illumination; and, iii) generalizable to multiple species. Using as starting point a fully convolutional network (FCN) \cite{Chen2016} pre-trained on a large multi-class dataset, we describe an effective fine-tuning procedure that adapts this model for fine pixel-wise flower segmentation. Our final method evaluates in less than $50$ seconds high-resolution images covering each a full tree. Although the task comparison is not one-to-one, human workers may need on average up to $50$ minutes to count the number of flowers per tree. 
    
    \item A feasible procedure for evaluating high-resolution images with deep FCNs on commercial GPUs. Fully convolutional computations require GPU memory space that exponentially increases according to image resolution. We employ an image partitioning mechanism with partially overlapping windows, which reduces artifacts introduced by artificial boundaries when evaluating disjoint image regions.
    
    \item Release of an annotated dataset with pixel-accurate labels for flower segmentation on high resolution images \cite{Tabb2018Datasets}. We believe this can greatly benefit the community, since this is a very time consuming yet critical task for both training and evaluation of segmentation models. 

\end{itemize}

\section{\uppercase{Related Work}}
Previous attempts at automating bloom intensity estimation were mostly based on color thresholding, such as the works described in \cite{Aggelopoulou2011color,Hocevar2014color} and \cite{Thorp2011}. Despite differences in terms of color-space used for analysis (e.g. HSL and RGB), all these methods fail when applied in uncontrolled environments. Apart from size filtering, no morphological feature is taken into account, such that thresholding parameters have to be adjusted in case of changes in illumination, camera position or flowering density. Even strategies using aerial multispectral images such as   \cite{Horton2017PeachAir} also rely solely on color information for image processing.

Our previous work in \cite{Dias2018} introduced a novel approach for apple flower detection that relies on a fine-tuned Clarifai CNN \cite{Zeiler2014clarifai} to classify individual superpixels composing an image. That method highly outperformed color-based approaches, especially in terms of generalization to datasets composed of different flower species and acquired in uncontrolled environments. However, existing superpixel algorithms rely solely on local context information, representing the main source of imprecisions in scenarios where flowers and the surrounding background present similar colors.

While early attempts for autonomous fruit detection also relied on hand-engineered features (e.g. color, texture, shape) \cite{Gongal2015}, recent works have been exploring more advanced computer vision techniques. One example is the work of Hung et al. \cite{Hung2013Fruit}, which combines sparse autoencoders \cite{Guo2016cnn} and support vector machines (SVM) for segmenting leaves, almonds, trunks, ground and sky. The approaches described by Bargoti and Underwood in \cite{Bargoti2016Fruit} and Chen et al. in \cite{Chen2017} for fruit detection share some similarities with our method for flower segmentation. In \cite{Bargoti2016Fruit}, the authors introduce a Faster R-CNN trained for the detection of mangoes, almonds and apple fruits on trees. The method introduced in \cite{Chen2017} for counting apples and oranges employs a fully convolutional network (FCN) to perform fruit segmentation and a convolutional network to estimate fruit count.

End-to-end fully convolutional networks \cite{Long2015fcn} have been replacing traditional fully connected architectures for image segmentation tasks \cite{Garcia2017rev}. Conventional architectures such as the \textit{Alexnet} \cite{Krizhevsky2012alexnet} and VGG \cite{Simonyan2014vgg} networks are very effective for image classification but provide coarse outputs for image segmentation tasks. This is a consequence of the image downsampling introduced by the \textit{max-pooling} and \textit{striding} operations performed by these networks, which allow the extraction of learned hierarchical features at the cost of pixel-level precision \cite{Chen2016}.

Different strategies have been proposed to alleviate the effects of downsampling \cite{Garcia2017rev}, including the use of deconvolution layers \cite{Long2015fcn,Noh2015deconv}, and encoder-decoder architectures with skip layer connections \cite{Kendall2015segnet,Hariharan2015hyper}. The DeepLab model introduced in \cite{Chen2016} is one of the most successful approaches for semantic image segmentation using deep learning. By combining the ResNet-101 \cite{He2016resnet} model with \textit{atrous} convolutions and spatial pyramid pooling, it significantly reduces the downsampling rate and achieves state-of-the-art performance in challenging semantic segmentation datasets such as the PASCAL VOC \cite{Everingham2015pascal} and COCO \cite{Lin2014coco}.

In addition to the changes in CNN architecture, the authors of DeepLab also employ the dense CRF model described in \cite{Krahenbuhl2012} to produce fine-grained segmentations. Although providing visually appealing segmentations, this refinement model relies on parameters that have to be optimized by means of supervised grid-search. In \cite{Dias2018rgr}, we introduced a generic post-processing module that can be coupled to the output of any CNN to refine segmentations without the need for dataset-specific tuning. Called region growing refinement (RGR), this algorithm uses the score maps available from the CNN to divide the image into regions of high confidence background, high confidence object and uncertainty region. By means of appearance-based region growing, pixels within the uncertainty region are classified based on initial seeds randomly sampled from the high confidence regions.

\begin{figure*}[t]
  \centering
  \includegraphics[trim={0cm 0.1cm 0 0.1cm},clip,width=0.95\textwidth]{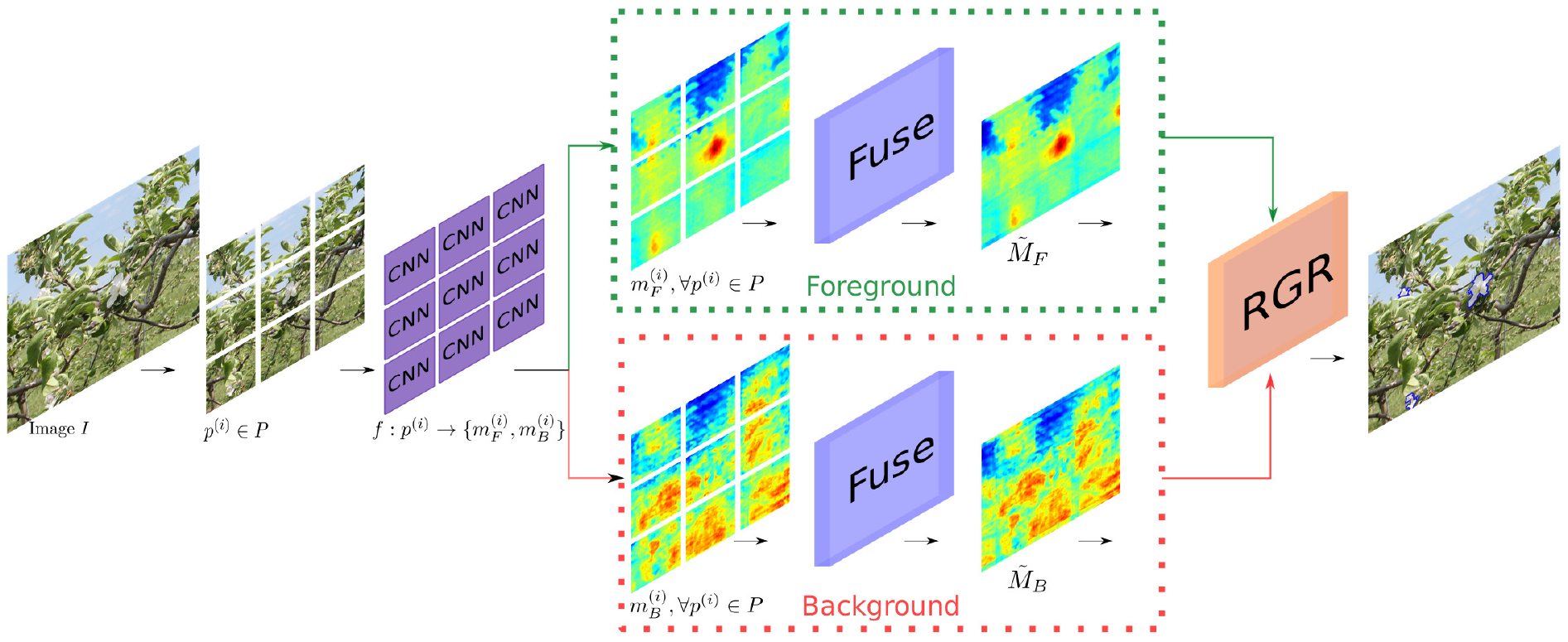}
    \caption{\textbf{Best viewed in color.} Diagram illustrating the sequence of tasks performed by the proposed method for flower detection. Each task and its corresponding output (shown below the arrows) are described in Algorithm \ref{alg:pseudocode}. In the heatmaps, blue is associated with lower scores, while higher scores are illustrated with red.}
    \label{fig:diagram}
\end{figure*}

\section{\uppercase{Our Approach}}
In this section, we first describe the pre-training and fine-tuning procedures carried out to obtain a CNN highly sensitive to flowers. Subsequently, we describe the sequence of operations that our pipeline performs to segment flowers in an image.

\subsection{Network training}

One of the largest datasets available for semantic segmentation, the COCO dataset \cite{Lin2014coco} was recently augmented by Caesar et al. \cite{Caesar2017stuff} into the COCO-Stuff dataset. This dataset includes pixel-level annotations of classes such as \textit{grass, leaves, tree} and \textit{flowers}, which are relevant for our application. In the same work, the authors also discuss the performance of modern semantic segmentation methods on COCO-Stuff, with a DeepLab-based model outperforming the standard FCN. Thus, we opted for the publicly available DeepLab-ResNet model pre-trained on the COCO-Stuff dataset as the starting point for our pipeline. Rather than fine-tuning the dense CRF model used in the original DeepLab work, we opt for the generic RGR algorithm as a post-processing module to obtain fine-grained segmentations.

The base model was originally designed for segmentation within the $172$ COCO-Stuff classes. To adapt its architecture for our binary flower segmentation task, we perform procedures known as \textit{network surgery} and \textit{fine-tuning} \cite{Girshick2014rcnn}. The surgery procedure is analogous to the pruning of undesired branches in trees: out of the original $172$ classification branches, we preserve only the weights and connections responsible for the segmentation of classes of interest.

We considered first an architecture preserving only the \textit{flower}  classification branch, followed by a sigmoid classification unit. However, without the normalization induced by the model's original softmax layer, the scores generated by the transferred \textit{flower} branch are unbounded and the final sigmoid easily saturates. To alleviate the learning difficulties caused by such a poor initialization, we opted for tuning a model with two-branches, under the hypothesis that a second branch would allow the network to learn a background representation that properly normalizes the predictions generated by the foreground (\textit{flower}) branch.

We have observed experimentally that nearby leaves represent one of the main sources of misclassification for flower segmentation. Moreover, predictions for the class \textit{leaf} presented the highest activations when applying the pre-trained model to our training dataset. For these reasons, we opt for this branch together with the one associated with \textit{flowers} to initialize our two-branch flower segmentation network. 

The adapted architecture was then fine-tuned using the training set described in Section \ref{sec:datasets}, which contains $100$ images of apple trees. For our experiments, the procedure was carried out for $10,000$ iterations using the Caffe framework \cite{Jia2014a}, with an initial learning rate of $10^{-4}$ that polynomially decays according to $10^{-4}\times(1 - i/10000)^{0.9}$, where $i$ is the iteration number. Aiming at scale robustness, our fine-tuning procedure employs the same strategy used for model pre-training, where each training portrait is evaluated at ($0.5$, $0.75$, $1.0$, $1.25$, $1.5$) times its original resolution.

While the validation set has pixel-accurate annotations obtained using the procedure described in Section \ref{sec:datasets}, the training set was annotated using the less precise but quicker superpixel-based procedure described in our previous work \cite{Dias2018}. Less than $5\%$ of the total image areas in this dataset contain flowers. To compensate for this imbalance, we augmented portraits containing flowers by mirroring them with respect to vertical and horizontal axes. Following the original network parameterization, we split the $100$ training images into portraits of $321\times321$ pixels, corresponding to a total of $52,644$ training portraits after augmentation.

\begin{algorithm}
  \caption{Proposed approach for flower detection}
  \label{alg:pseudocode}
  \begin{algorithmic}[1]
      \Require{Image $I$.}
      \Ensure{Estimated flower segmentation map $\hat{Y}$ of image $I$.} 
      \State Sliding window: divide $I$ into a set of $n$ portraits $P$. \label{ln:sw}
      \For {each portrait $p^{(i)} \in P$}
            \State \parbox[t]{\dimexpr\linewidth-\algorithmicindent}{Compute scoremaps $m_B^{(i)}$ and 	$m_F^{(i)}$ using the fine-tuned CNN}\vspace{5pt}
      \EndFor
      \State Obtain $M_B$ and $M_F$ by fusing $m_B^{(i)}$ and $m_F^{(i)}$ ($i=1,\ldots,n$), respectively according to Eq. \ref{eq:fusion}. \label{ln:fuse}
      \State Normalize $M_B$ and $M_F$ into $\Tilde{M}_B$ and $\Tilde{M}_F$, respectively according to Eq. \ref{eq:softmax}. \label{ln:soft}      
      \State Generate $\hat{Y}$ by applying RGR to $\Tilde{M}_B$ and $\Tilde{M}_F$.
  \end{algorithmic}
\end{algorithm}

\subsection{Segmentation pipeline}
The method we propose for fruit flower segmentation consists of three main operations: 1) divide a high resolution image into smaller patches, in a sliding window manner; 2) evaluate each patch using our fine-tuned CNN; 3) apply the refinement algorithm on the obtained scoremaps to compute the final segmentation mask. These steps are described in detail below. In our description, we make reference to Algorithm \ref{alg:pseudocode} and Figure \ref{fig:diagram}.

\begin{figure}[h]
	\centering	      
	\includegraphics[width=0.48\textwidth]{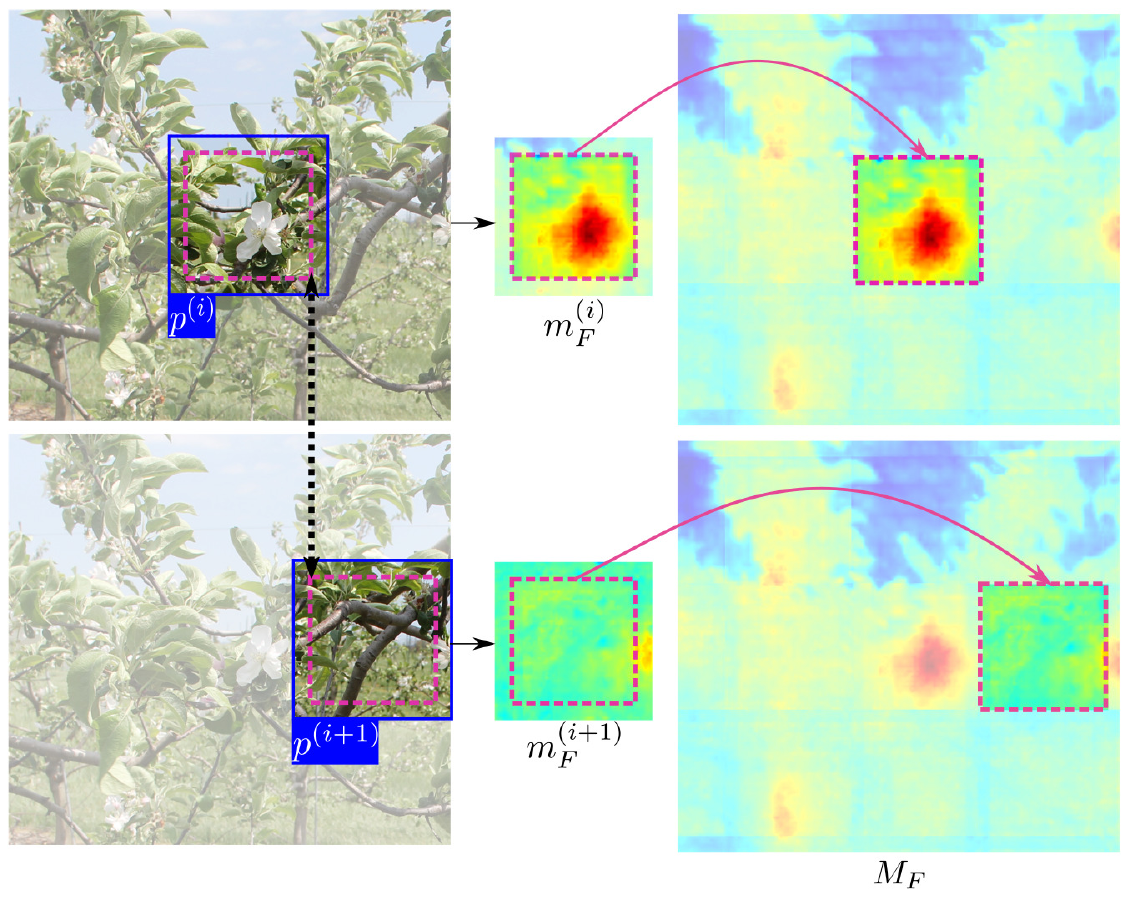}	
    \caption{\textbf{Best viewed in color.} Illustration of the sliding window and subsequent fusion process that comprise our segmentation pipeline. Each portrait overlaps a certain area of its neighbors, which is discarded during fusion to avoid artifacts caused by artificial boundaries.}    
    \label{fig:pad}
\end{figure}

\textit{1) Step 1 - Sliding window:} As mentioned above, the adopted CNN architecture either crops or resizes input images to $321\times321$ portraits. Since our datasets are composed of images with resolution ranging from $2704\times1520$ to $5184\times3456$ pixels (see Section \ref{sec:datasets}), we emulate a sliding window approach to avoid resampling artifacts. More specifically, we split each input image $I$ into a set $P$ of $n$ portraits $p^{(i)} \in P$. Each portrait is $321\times321$ pixels large, i.e. $p^{(i)} \in \mathbb{R}^{r\times r}$ with $r = 321$. Cropping non-overlapping portraits from the original image introduces artificial boundaries that compromise the detection quality. For this reason, in our approach each portrait overlaps a percentage $s$ of the area of each immediate neighbor. For our experiments, we adopted $s = 10\%$. When the scoremaps are fused, the results corresponding to the overlapping pixels are discarded. Figure \ref{fig:pad} illustrates this process for a pair of subsequent portraits. The scores obtained for each portrait are depicted as a heatmap, where blue is associated with lower scores and higher scores are illustrated with red.

\begin{table*}[t!]
\centering
\caption{Datasets specifications.}
\label{tab:datasets}
\begin{tabular}{lcccccc}
\hline
\multicolumn{1}{c}{\textbf{Dataset}} & \textbf{\begin{tabular}[c]{@{}c@{}}No. images\end{tabular}} & \textbf{Weather} & \textbf{\begin{tabular}[c]{@{}c@{}}Background \\ panel\end{tabular}} & \textbf{\begin{tabular}[c]{@{}c@{}}Camera\\ model\end{tabular}} & \textbf{Resolution} & \textbf{\begin{tabular}[c]{@{}c@{}}Camera \\ support\end{tabular}} \\ \hline\hline
\textit{AppleA} & \begin{tabular}[c]{@{}c@{}}$100$ (train) + $30$ (val)\end{tabular} & Sunny & No & Canon EOS 60D & $5184\times3456$ & Hand-held \\
\textit{AppleB} & $18$ & Sunny & Yes & GoPro HERO5 & $2704\times1520$ & Utility vehicle \\
\textit{Peach} & $24$ & Overcast & No & GoPro HERO5 & $2704\times1520$ & Hand-held \\
\textit{Pear} & $18$ & Overcast & No & GoPro HERO5 & $2704\times1520$ & Hand-held \\ \hline
\end{tabular}
\end{table*}

\begin{table}[]
\centering
\caption{HSV statistics of images composing each dataset.}
\label{tab:imghsv}
\begin{tabular}{lccccccc}
\cline{2-7} 
\multicolumn{1}{c}{} & \multicolumn{2}{c}{\textbf{H{ [}$\mathbf{0-360^\circ}${]}}} & \multicolumn{2}{c}{\textbf{S{ [}\%{]}}} & \multicolumn{2}{c}{\textbf{V{ [}\%{]}}}\tabularnewline
\hline 
Dataset  &  \textit{$\mu_{H}$}  & \textit{$IQR_{H}$}  & \textit{$\mu_{S}$}  & \textit{$IQR_{S}$}  & \textit{$\mu_{V}$}  & \textit{$IQR_{V}$} \tabularnewline
\hline 
\hline 
\textit{AppleA} & $74.6$ & $49.3$ & $32.9$ & $24.3$ & $53.7$  & $30.2$ \tabularnewline
\textit{AppleB} & $219.6$ & $21.1$ & $88.6$ & $44.3$ & $47.1$ & $16.9$ \tabularnewline
\textit{Peach} & $223.8$ & $199.9$ & $11.8$ & $20.7$ & $42.3$ & $46.6$ \tabularnewline
\textit{Pear} & $85.9$ & $178.8$ & $16.4$ & $23.4$ & $42.4$ & $20.8$ \tabularnewline
\hline 
\end{tabular}
\end{table}

\begin{table}[]
\centering
\caption{HSV statistics of flowers composing each dataset.}
\label{tab:flhsv}
\begin{tabular}{lcccccc}
\cline{2-7} 
\multicolumn{1}{c}{} & \multicolumn{2}{c}{\textbf{H{ [}$\mathbf{0-360^\circ}${]}}} & \multicolumn{2}{c}{\textbf{S{ [}\%{]}}} & \multicolumn{2}{c}{\textbf{V{ [}\%{]}}}\tabularnewline
\hline 
Dataset  & \textit{$\mu_{H}$}  & \textit{$IQR_{H}$}  & \textit{$\mu_{S}$}  & \textit{$IQR_{S}$}  & \textit{$\mu_{V}$}  & \textit{$IQR_{V}$} \tabularnewline
\hline 
\hline 
\textit{AppleA} & $136.6$ & $205.5$ & $6.3$ & $9.8$ & $77.3$ & $24.3$  \tabularnewline
\textit{AppleB} & $56.3$ & $80.2$ & $7.5$  & $9.8$  & $86.7$ & $23.1$  \tabularnewline
\textit{Peach} & $325.2$ & $26.7$ & $21.2$  & $13.3$  & $50.2$ & $13.7$ \tabularnewline
\textit{Pear} & $215.4$ & $173.2$ & $5.9$  & $5.9$  & $84.7$ & $22.4$  \tabularnewline
\hline 
\end{tabular}
\end{table}

\begin{figure}[h]
	\centering	      
    \subfloat[$m_B^{(i)}$]{\includegraphics[width=0.23\linewidth]{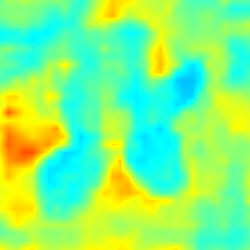}}\hspace{0.1mm}
	\subfloat[$m_F^{(i)}$]{\includegraphics[width=0.23\linewidth]{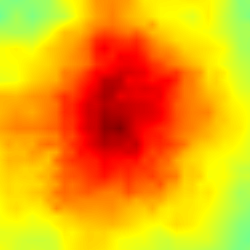}}\hspace{0.1mm}	
	\subfloat[Coarse \newline segmentation]{\includegraphics[width=0.23\linewidth]{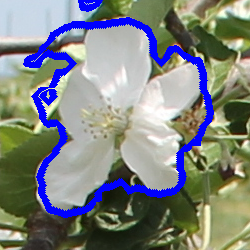}}\hspace{0.1mm}	
	\subfloat[Refined \newline segmentation]{\includegraphics[width=0.23\linewidth]{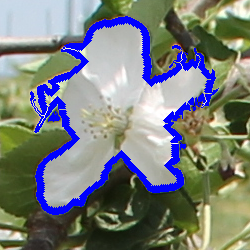}}
    \caption{\textbf{Best viewed in color.} Example of segmentation refinement for a given pair of scoremaps. a) Background scoremap $m_B^{(i)}$. b) Foreground scoremap $m_F^{(i)}$. c) Coarse segmentation by direct thresholding of the scoremaps. d) Refined segmentation using RGR. }    
    \label{fig:refine}
\end{figure}

\textit{2) Step 2 - CNN prediction:} We evaluate in parallel each portrait $p^{(i)}$  with our fine-tuned network for flower identification. The CNN is equivalent to a function $f$ 
\begin{equation}
f: p^{(i)} \rightarrow \{m_F^{(i)},m_B^{(i)}\},
\end{equation}
which maps each input $p^{(i)}$ into two pixel-dense scoremaps: $m_F^{(i)} \in \mathbb{R}^{r\times r}$ represents the pixel-wise likelihood that pixels in $p^{(i)}$ belong to the foreground (i.e., flower), while $m_B^{(i)} \in \mathbb{R}^{r\times r}$ corresponds to the pixel-wise background likelihood. The heatmaps in Figures \ref{fig:refine}(a) and (b) are examples of scoremaps computed for a given portrait.

\textit{3) Step 3 - Fusion and refinement:} After evaluating each portrait, we generate two global scoremaps $M_B$ and $M_F$ by combining the predictions obtained for all $p^{(i)} \in P$. Let $c^{(i)}$ represent the pixel-coordinates of $p^{(i)}$ in $I$ after discarding the padding pixels. The fusion procedure is defined as
\begin{equation}
\label{eq:fusion}
\forall p^{(i)} \in P, \quad M_{F,B}(c^{(i)})=m_{F,B}^{(i)}, 
\end{equation}
such that both scoremaps $M_B$ and $M_F$ have the same resolution as $I$. As illustrated in Figure \ref{fig:pad}, the padded areas of $m_{F,B}^{(i)}$ (outside the red box) are discarded during fusion. For every pixel in the image, a single prediction score is obtained from exactly one portrait, such that artifacts introduced by artificial boundaries are avoided.

After fusion, the scoremaps $M_B$ and $M_F$ are normalized into scoremaps $\Tilde{M}_B$ and $\Tilde{M}_F$ using a softmax function
\begin{equation}
\label{eq:softmax}
\Tilde{M}_{F,B}(q_j)=\frac{\exp(M_{F,B}(q_j))}{\exp(M_{B}(q_j))+\exp(M_{F}(q_j))}, 
\end{equation}
where $q_j$ is the $j$-th pixel in the input image $I$. With this formulation, for each pixel $q_j$ the scores $\Tilde{M}_B(q_j)$ and $\Tilde{M}_F(q_j)$ add to one, i.e. they correspond to the probability that $q_j$ belongs to the corresponding class.

As Figure \ref{fig:refine}(c) shows, the predictions obtained directly from the CNN are coarse in terms of adherence to actual flower boundaries. Therefore, rather than directly thresholding $\Tilde{M}_F$, this scoremap and the image $I$ are fed to the RGR refinement module described in \cite{Dias2018rgr}. For our application, the refinement algorithm relies on two high-confidence classification regions $R_F$ and $R_B$ defined according to
\begin{alignat}{3}
R_{F,B}&=\left\{ q_{j}|\Tilde{M}_{F,B}(q_{j})>\tau_{F,B}\right\}, \label{eq:ru}
\end{alignat}
where $\tau_B$ and $\tau_F$ are the high-confidence background and foreground thresholds. Using the high-confidence regions as starting points, the RGR algorithm performs multiple Monte Carlo region growing steps that groups similar pixels into clusters. Afterwards, it performs majority voting to classify each cluster according to the presence of flowers. Each pixel $q_j$ within a cluster contributes with a positive vote if its score $\Tilde{M}_F(q_j)$ is larger than a threshold $\tau_0$. As detailed in Section \ref{sec:results}, this parameter can be empirically tuned according to the dataset under consideration. Based on a grid-search optimization on our training dataset, we selected $\tau_0=0.3$ for all our experiments and fixed $\tau_B=0.1$ and $\tau_F=1.25\times\tau_0$.

\section{\uppercase{Datasets}}
\label{sec:datasets}

We evaluate our method on four datasets that we created and made publicly available: \textit{AppleA}, \textit{AppleB}, \textit{Peach}, \textit{Pear} \cite{Tabb2018Datasets}. As summarized in Table \ref{tab:datasets}, images from different fruit flower species were collected in diverse uncontrolled environments and under different angles of capture. 

\begin{figure}[b]
	\centering	      
	\includegraphics[trim={0cm 0.3cm 0 4cm},clip,width=0.45\textwidth]{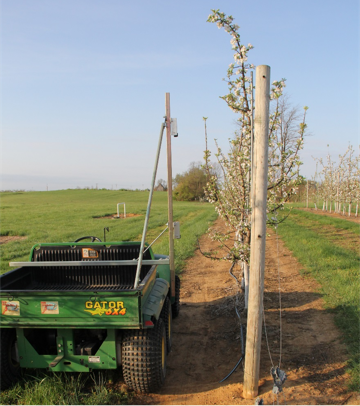} 
    \caption{\textbf{Best viewed in color.} Utility vehicle used for imaging. For the \textit{AppleB} dataset, this vehicle was used in conjunction with a background panel.}
   \label{fig:gator}
\end{figure}

Both datasets \textit{AppleA} and \textit{AppleB} are composed of images of apple trees, which were collected in a USDA orchard on a sunny day. In both datasets, the trees are supported with trellises and planted in rows. \textit{AppleA} is a collection of $147$ images acquired using a hand-held camera. From this total, we randomly selected $100$ images to build the training set used to train the CNN. Out of the remaining $47$ images, $30$ were randomly selected to compose the testing set for which we report results in Section \ref{sec:results}.

This dataset contains flowers that greatly vary in terms of size, cluttering, occlusion by leaves and branches. Flowers composing its images have an average area of $10,730$ pixels, but with a standard deviation of $17,150$ pixels. On average, flowers compose only $2.5\%$ of the total image area within this dataset, which is otherwise vastly occupied by leaves.

Differently from \textit{AppleA}, for the \textit{AppleB} dataset, a utility vehicle equipped with a background unit was used for imaging, such that trees in other rows are not visible in the images. Figure \ref{fig:gator} illustrates the utility vehicle used for image acquisition, and Figures \ref{fig:appleA} and \ref{fig:appleB} illustrate the differences between datasets \textit{AppleA} and \textit{AppleB}.

The \textit{Peach} and \textit{Pear} datasets differ both in terms of species and acquisition conditions, therefore representing adequate scenarios for evaluating the generalization capabilities of the proposed method. Both datasets contain images acquired on an overcast day and without a background unit. Compared to the \textit{AppleA} dataset, images composing these datasets present significantly lower saturation and value means. Tables \ref{tab:imghsv} and \ref{tab:flhsv} summarize the differences among datasets in terms of the statistics of the HSV color components, where $\mu$ stands for mean values and $IQR$ for interquartile ranges.

Regarding the flower characteristics, apple blossoms are typically white, with hue components spread in the whole spectrum (high $IQR_H$) and low saturation mean. Flowers composing the \textit{AppleB} dataset present higher brightness ($\mu_V$), while peach flowers show a pink hue centered on $\mu_H=325^\circ$, with higher saturation and lower value means. Moreover, pear flowers are slightly different in terms of color (greener) and morphology, as illustrated in Figure \ref{fig:pear}.

\subsection{Labeling}
Image annotation for segmentation tasks is a laborious and time-consuming activity. Labels must be accurate at pixel-level, otherwise both supervised training and the evaluation of segmentation techniques are compromised. Most existing annotation tools rely on approximating segmentations as polygons, which provide ground truth images that frequently lack accurate adherence to real object boundaries \cite{Dias2018rgr}. 
\begin{figure}[h]
	\centering	      
	\includegraphics[trim={3cm 0cm 0 0cm},clip,width=0.49\linewidth]{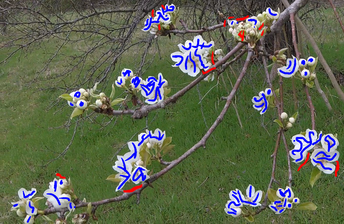}
    \includegraphics[trim={3cm 0cm 0 0cm},clip,width=0.49\linewidth]{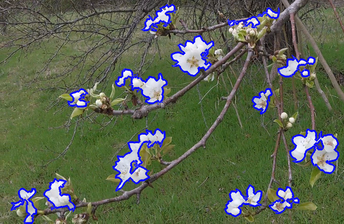}
    \caption{\textbf{Best viewed in color.} Example of ground truth obtained from freehand annotations. \textit{Left:} positive examples are annotated in blue, while hard negatives are indicated in red. \textit{Right:} segmentation obtained after RGR refinement.}    
    \label{fig:label}
\end{figure}

We opted for a labeling procedure that combines freehand annotations and RGR refinement \cite{Dias2018rgr}. Using a tablet, the user draws traces on regions of the image that contain flowers, indicating as well hard negative examples when necessary. These traces indicate high-confidence segmentation points, which are used as reference by RGR to segment the remaining parts of the image. Figure \ref{fig:label} shows an example of a ground truth segmentation obtained using this procedure\footnote{We will make the annotation tool publicly available as future work.}.

\section{\uppercase{Experiments and results}}
\label{sec:results}
We aim at a method capable of accurate multi-species flower detection, regardless of image acquisition conditions and without the need for dataset-specific training or pre-processing. To verify that our method satisfies all these requirements, we performed experiments on the four different datasets described in Section \ref{sec:datasets} while only using the \textit{AppleA} dataset for training. 

We adopt as the main baseline our previous model described in \cite{Dias2018}, which highly outperformed existing methods by employing the Clarifai CNN architecture to classify individual superpixels. We therefore refer to that model as \textsc{Sppx+Clarifai} and to our new method as \textsc{DeepLab+RGR}. We also compare our results against a \textsc{HSV-based} method \cite{Hocevar2014color} that segments images based only on HSV color information and size filtering according to threshold values optimized using grid-search.

\setcounter{figure}{5}
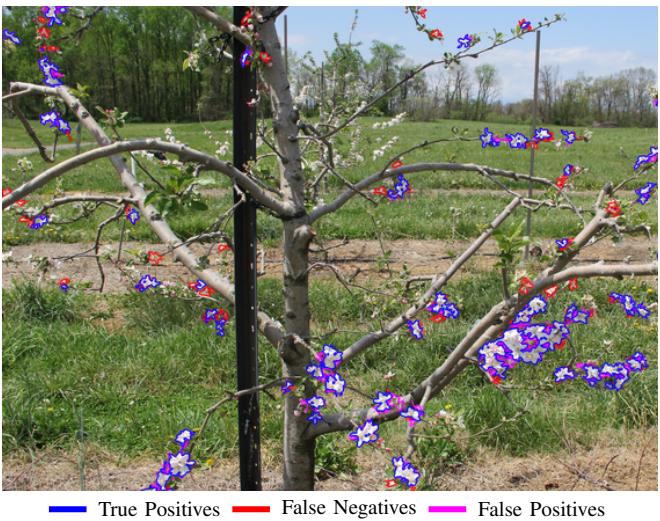
\begin{figure}[h]
	\centering	      
	\includegraphics[trim={0cm 2.5cm 7.5cm 1cm},clip,width=0.48\textwidth]{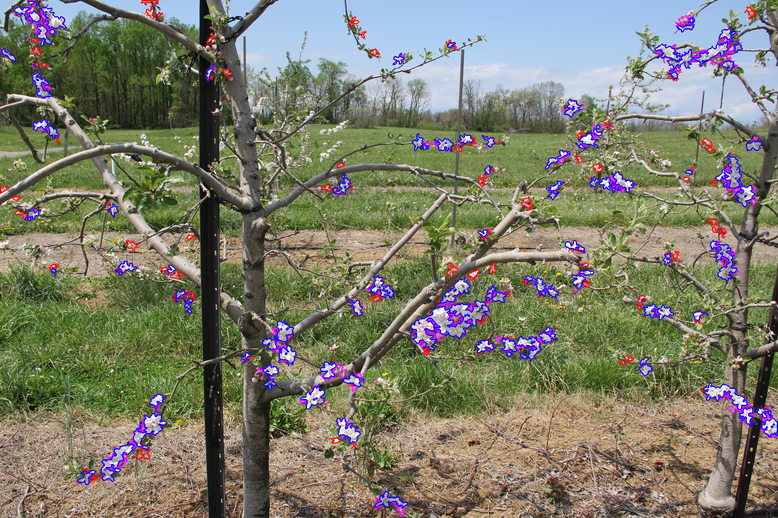}
    \begin{tikzpicture}
	\draw [line width=0.8mm, blue,left] (0,0) -- (.5,0) node [right,color=black] (text1) {\footnotesize{True Positives}};;
	\draw [line width=0.8mm, red] (text1.east) -- ([xshift=5mm]text1.east) node [right,color=black] (text2) {\footnotesize{False Negatives}};;
	\draw [line width=0.8mm, colorPink] (text2.east) -- ([xshift=5mm]text2.east) node [right,color=black] (text3) {\footnotesize{False Positives}};;
    \end{tikzpicture}       
    \caption{\textbf{Best viewed in color.} Examples of flower detection in one image composing the \textit{AppleA} dataset.\vspace{0.0cm}}
    \label{fig:appleA}
\end{figure}

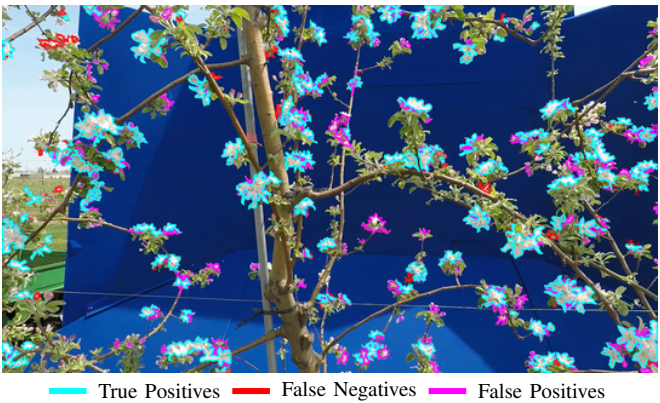
\begin{figure}[h]
	\centering	      
	\includegraphics[width=0.48\textwidth]{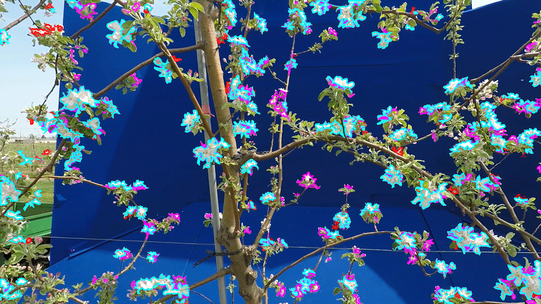}
    \begin{tikzpicture}
	\draw [line width=0.8mm, colorCyan,left] (0,0) -- (.5,0) node [right,color=black] (text1) {\footnotesize{True Positives}};;
	\draw [line width=0.8mm, red] (text1.east) -- ([xshift=5mm]text1.east) node [right,color=black] (text2) {\footnotesize{False Negatives}};;
	\draw [line width=0.8mm, colorPink] (text2.east) -- ([xshift=5mm]text2.east) node [right,color=black] (text3) {\footnotesize{False Positives}};;
    \end{tikzpicture}       
    \caption{\textbf{Best viewed in color.} Examples of flower detection in one image composing the \textit{AppleB} dataset.}
    \label{fig:appleB}
\end{figure}

All three methods were tuned using the \textit{AppleA} training dataset, with differences in the pipeline for transfer learning. For the three unseen datasets, the \textsc{Sppx+Clarifai} relies on a pre-processing step that enhances contrast and removes the different backgrounds present in the images. Our new method \textsc{DeepLab+RGR} does not require any pre-processing. Instead, it employs the same pipeline regardless of the dataset, requiring only adjustments in portrait size. As summarized in Table \ref{tab:datasets}, images composing the \textit{AppleA} dataset have resolution $4.3\times$ larger than images in the other three datasets. Thus, we split images in these datasets into portraits of $155\times155$ pixels, rather than the $321\times321$ pixels portraits used for \textit{AppleA}.

\setcounter{figure}{7}
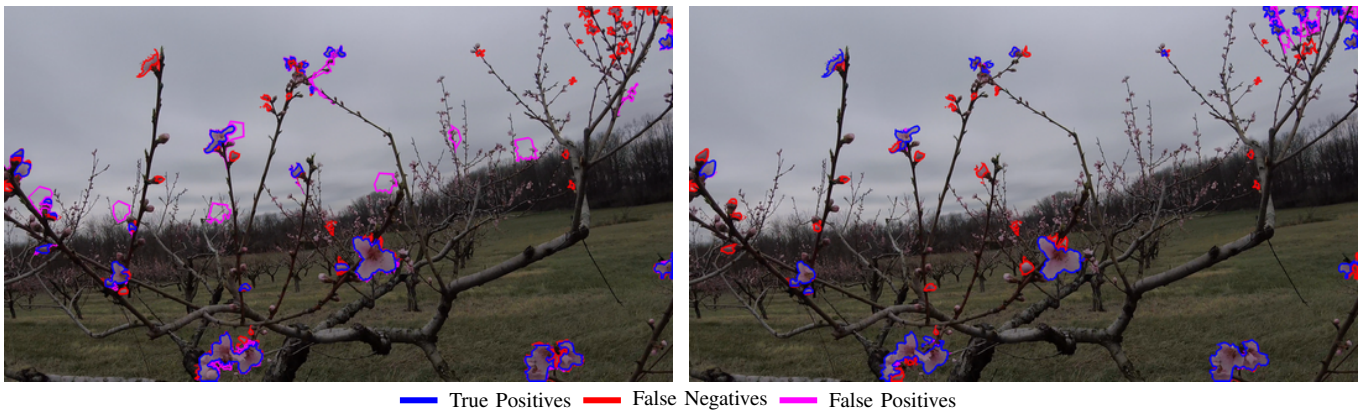
\begin{figure*}[!t]
	\centering	      
	\includegraphics[width=0.49\textwidth]{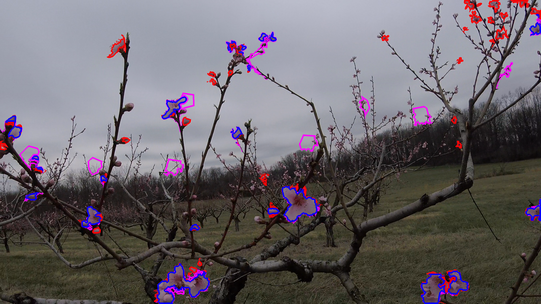}\hspace{1mm}
    \includegraphics[width=0.49\textwidth]{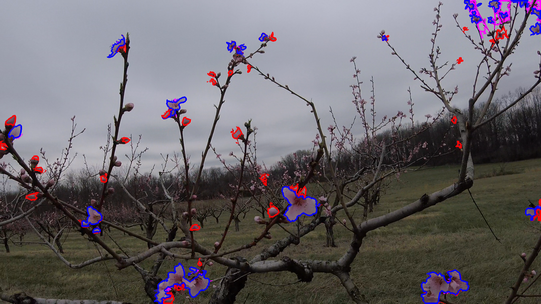}
    \begin{tikzpicture}
	\draw [line width=0.8mm, blue,left] (0,0) -- (.5,0) node [right,color=black] (text1) {\footnotesize{True Positives}};;
	\draw [line width=0.8mm, red] (text1.east) -- ([xshift=5mm]text1.east) node [right,color=black] (text2) {\footnotesize{False Negatives}};;
	\draw [line width=0.8mm, colorPink] (text2.east) -- ([xshift=5mm]text2.east) node [right,color=black] (text3) {\footnotesize{False Positives}};;
    \end{tikzpicture}     
    \caption{\textbf{Best viewed in color.} Examples of flower detection in one image composing the \textit{Peach} dataset. \textit{Left:} detections provided by the \textsc{Sppx+Clarifai} method. \textit{Right:} detections obtained with our new \textsc{DeepLab+RGR} method.}    
    \label{fig:peach}
\end{figure*}

The quantitative analysis of segmentation accuracy relies on precision, recall, $F_1$ and intersection-over-union (IoU) metrics \cite{Everingham2015pascal} computed at pixel-level, instead of the superpixel-wise metrics used in our previous work. Table \ref{tab:results} summarizes the results obtained by each method on the different datasets. 

\begin{table}[h]
\setlength\tabcolsep{5pt}
\centering
\caption{Summary of results obtained for each method.}\label{tab:results}
\begin{tabular}{llcccc}
\hline
\textit{} &  & \textbf{IoU} & {$\mathbf{F_1}$} & \textbf{Recall} & \textbf{Precision} \\ \hline\hline
\multirow{3}{*}{\textit{AppleA}} & \textsc{HSV-based} & $28.0\%$ & $43.7\%$ & $56.5\%$ & $35.7\%$ \\
 & \textsc{Sppx+Clarifai} & $51.3\%$ & $67.8\%$ & $73.2\%$ & $63.1\%$ \\
 & \textsc{DeepLab+RGR} & $\mathbf{71.4\%}$ & $\mathbf{83.3\%}$ & $\mathbf{87.7\%}$ & $\mathbf{79.4\%}$ \\ \hline
\multirow{3}{*}{\textit{AppleB}} & \textsc{HSV-based}& $49.3\%$ & $66.0\%$ & $58.9\%$ & $75.1\%$ \\
& \textsc{Sppx+Clarifai} & $50.6\%$ & $67.2\%$ & $68.4\%$ & $66.1\%$ \\
& \textsc{DeepLab+RGR} & $\mathbf{63.0\%}$ & $\mathbf{77.3\%}$ & $\mathbf{91.2\%}$ & $\mathbf{67.1\%}$ \\ \hline
\multirow{3}{*}{\textit{Peach}} & \textsc{HSV-based} & $0.1\%$ & $1.4\%$ & $1.4\%$ & $1.6\%$ \\
 & \textsc{Sppx+Clarifai} & $49.1\%$ & $67.2\%$ & $71.3\%$ & $61.2\%$ \\
 & \textsc{DeepLab+RGR} & $\mathbf{59.0\%}$ & $\mathbf{74.2\%}$ & $\mathbf{64.8\%}$ & $\mathbf{86.8\%}$ \\ \hline
\multirow{3}{*}{\textit{Pear}} & \textsc{HSV-based} & $39.7\%$ & $56.8\%$ & $65.6\%$ & $50.1\%$ \\
& \textsc{Sppx+Clarifai} & $40.5\%$ & $57.6\%$ & $49.6\%$ & $68.7\%$ \\
 & \textsc{DeepLab+RGR} & $\mathbf{75.4\%}$ & $\mathbf{86.0\%}$ & $\mathbf{79.2\%}$ & $\mathbf{94.1\%}$ \\ \hline 
\end{tabular}
\end{table}

Our new model outperforms the baseline methods for all datasets evaluated, especially in terms of generalization to unseen datasets. By combining a deeper CNN architecture and the RGR refinement module, \textsc{DeepLab+RGR} improves both prediction and recall rates in the validation \textit{AppleA} set by more than $15\%$. Figure \ref{fig:appleA} provides a qualitative example of flower detection accuracy in this dataset.

As Figure \ref{fig:appleB} illustrates, images composing the \textit{AppleB} dataset present a higher number of flower buds and illumination changes, especially in terms of sunlight reflection by leaves. Despite the larger variance in comparison to the previous dataset, the performance obtained by \textsc{DeepLab+RGR} surpasses $77\%$ in terms of $F_1$. 

Results obtained for the \textit{Peach} dataset demonstrate the limitation of color-based methods and two important generalization characteristics of our model. The \textsc{HSV-based} method is incapable of detecting peach flowers, since their pink color is very different from the white apple blossoms used for training. On the other hand, our method presents $F_1$ near $75\%$, indicating that it can properly detect even flowers that differ to a great extent from apple flowers in terms of color. Moreover, images composing this dataset are characterized by a cloudy sky and hence poorer illumination. Most cases of false negatives correspond to flower buds, due to the lack of such examples in the training dataset. As illustrated in Figure \ref{fig:peach}, poor superpixel segmentation leads the \textsc{Sppx+Clarifai} approach to incorrectly classify parts of the sky as flowers. This problem is overcome by our new model, which greatly increases precision rates to above $80\%$. 
\setcounter{figure}{8}
\begin{figure}[!h]
	\centering	      
	\includegraphics[trim={0cm 2.9cm 5cm 0},clip,width=0.48\textwidth]{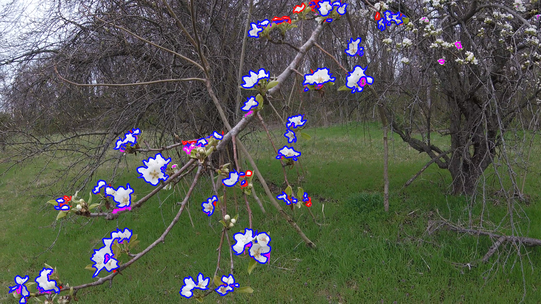}
    \begin{tikzpicture}
	\draw [line width=0.8mm, blue,left] (0,0) -- (.5,0) node [right,color=black] (text1) {\footnotesize{True Positives}};;
	\draw [line width=0.8mm, red] (text1.east) -- ([xshift=5mm]text1.east) node [right,color=black] (text2) {\footnotesize{False Negatives}};;
	\draw [line width=0.8mm, colorPink] (text2.east) -- ([xshift=5mm]text2.east) node [right,color=black] (text3) {\footnotesize{False Positives}};;
    \end{tikzpicture}       
    \caption{\textbf{Best viewed in color.} Examples of flower detection in one image composing the \textit{Pear} dataset.}
    \label{fig:pear}
\end{figure}
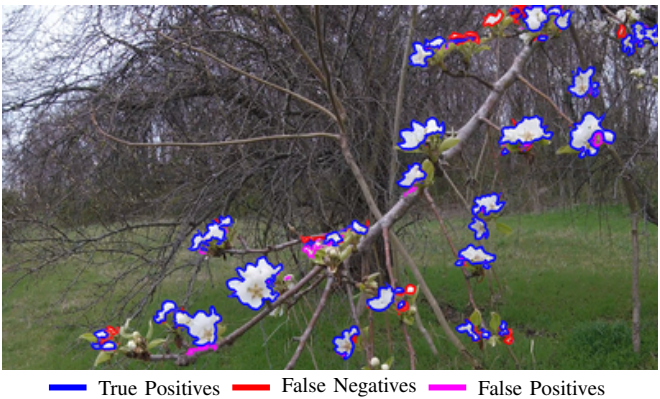

Furthermore, the high recall rate provided by \textsc{DeepLab+RGR} in the \textit{Pear} dataset demonstrates its robustness to slight variations in both flower morphology and color. As shown in Figure \ref{fig:pear}, similar to the \textit{Peach} dataset, these images also present a cloudy background. In addition to that, their background is characterized by a high level of clutter caused by the presence of a large number of branches. These high texture components compromise the background removal model used by \textsc{Sppx+Clarifai}. Still, the \textsc{DeepLab+RGR} method provides a very accurate detection of flowers, with precision above $90\%$.

\begin{figure}[h]
	\centering	      
	\includegraphics[width=0.49\textwidth]{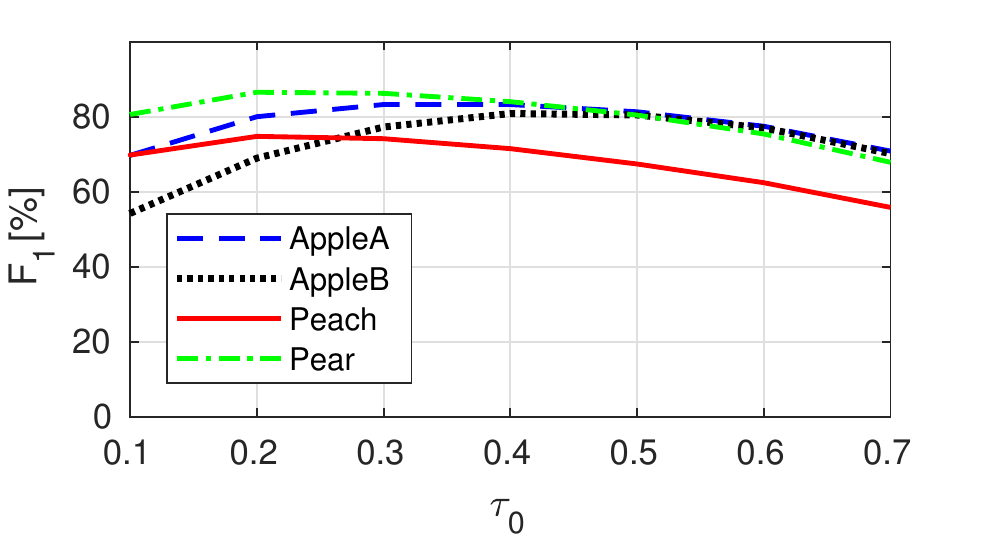}   
    \caption{Segmentation performance in terms of $F_1$ measure on each dataset according to the parameter $\tau_0$.}
    \label{fig:tauF}
\end{figure}

The results obtained by our method for \textit{AppleB}, \textit{Peach} and \textit{Pear} datasets can be further improved by adjusting the parameter $\tau_0$ used for final classification and refinement. As summarized in Figure \ref{fig:tauF}, increasing $\tau_0$ from $0.3$ to $0.5$ increases in $3\%$ the $F_1$ performance on \textit{AppleB}, reaching both recall and precision levels around $80\%$. For the \textit{Peach} dataset, decreasing $\tau_0$ to $0.2$ increases the recall rate to above $70\%$. Such adjustment can be carried out quickly through a simple interactive procedure, where $\tau_0$ is chosen according to its visual impact on the segmentation of a single image.

In terms of inference time, the current implementation of our algorithm on an Intel Xeon\texttrademark CPU E5-2620 v3 @ 2.40GHz (62GB) with a Quadro P6000 GPU requires on average $50$ seconds to evaluate each high-resolution image composing our datasets. Around $5$ seconds are required to save portraits as individual files and load their corresponding prediction scores, a process that can be simplified by generating portraits directly within the neural network framework.

\section{\uppercase{Conclusion}}
We have presented a novel automated approach for flower detection, which exploits state-of-the-art deep learning techniques for semantic image segmentation. The applicability of our method was demonstrated by its high flower segmentation accuracy across datasets that vary in terms of illumination conditions, background composition, image resolution, flower density and flower species. Without any supervised fine-tuning or image pre-processing, our model trained using only images of apple flowers succeeded in generalizing for peach and pear flowers, which are noticeably different in terms of color and morphology.

In the future, we intend to further improve the generalization capabilities of our model by training and evaluating it on multi-species flower datasets. We ultimately aim at a completely autonomous system capable of online bloom intensity estimation. The current implementation of our model can evaluate high-resolution images of complete trees an order of magnitude faster than human workers. While in this work we are not creating maps of flowers at the block level, this method will scale well for precision agricultural applications such as predicting thinning spray treatments and timing.







\bibliographystyle{IEEEtran}
\bibliography{refs}

\begin{thebibliography}{10}
\providecommand{\url}[1]{#1}
\csname url@rmstyle\endcsname
\providecommand{\newblock}{\relax}
\providecommand{\bibinfo}[2]{#2}
\providecommand\BIBentrySTDinterwordspacing{\spaceskip=0pt\relax}
\providecommand\BIBentryALTinterwordstretchfactor{4}
\providecommand\BIBentryALTinterwordspacing{\spaceskip=\fontdimen2\font plus
\BIBentryALTinterwordstretchfactor\fontdimen3\font minus
  \fontdimen4\font\relax}
\providecommand\BIBforeignlanguage[2]{{%
\expandafter\ifx\csname l@#1\endcsname\relax
\typeout{** WARNING: IEEEtran.bst: No hyphenation pattern has been}%
\typeout{** loaded for the language `#1'. Using the pattern for}%
\typeout{** the default language instead.}%
\else
\language=\csname l@#1\endcsname
\fi
#2}}

\bibitem{Forshey1986}
C.~Forshey, ``Chemical fruit thinning of apples,'' \emph{New York State
  Agricultural Experiment Station}, 1986.

\bibitem{Link2000}
H.~Link, ``Significance of flower and fruit thinning on fruit quality,''
  \emph{Plant growth regulation}, vol.~31, no. 1-2, pp. 17--26, 2000.

\bibitem{Gongal2016}
A.~Gongal, A.~Silwal, S.~Amatya, M.~Karkee, Q.~Zhang, and K.~Lewis, ``Apple
  crop-load estimation with over-the-row machine vision system,''
  \emph{Computers and Electronics in Agriculture}, vol. 120, pp. 26--35, 2016.

\bibitem{Zhang2002precision}
N.~Zhang, M.~Wang, and N.~Wang, ``Precision agriculture -- a worldwide
  overview,'' \emph{Computers and Electronics in Agriculture}, vol.~36, no.
  2-3, pp. 113--132, 2002.

\bibitem{Kapach2012}
K.~Kapach, E.~Barnea, R.~Mairon, Y.~Edan, and O.~Ben-Shahar, ``Computer vision
  for fruit harvesting robots--state of the art and challenges ahead,''
  \emph{International Journal of Computational Vision and Robotics}, vol.~3,
  no. 1-2, pp. 4--34, 2012.

\bibitem{Gongal2015}
A.~Gongal, S.~Amatya, M.~Karkee, Q.~Zhang, and K.~Lewis, ``Sensors and systems
  for fruit detection and localization: A review,'' \emph{Computers and
  Electronics in Agriculture}, vol. 116, pp. 8--19, 2015.

\bibitem{Guo2016cnn}
Y.~Guo, Y.~Liu, A.~Oerlemans, S.~Lao, S.~Wu, and M.~S. Lew, ``Deep learning for
  visual understanding: A review,'' \emph{Neurocomputing}, vol. 187, pp.
  27--48, 2016.

\bibitem{Chen2017}
S.~W. Chen, S.~S. Shivakumar, S.~Dcunha, J.~Das, E.~Okon, C.~Qu, C.~J. Taylor,
  and V.~Kumar, ``Counting apples and oranges with deep learning: a data-driven
  approach,'' \emph{IEEE Robotics and Automation Letters}, vol.~2, no.~2, pp.
  781--788, 2017.

\bibitem{Dyrmann2016crops}
M.~Dyrmann, A.~K. Mortensen, H.~S. Midtiby, and R.~N. J{\o}rgensen,
  ``Pixel-wise classification of weeds and crops in images by using a fully
  convolutional neural network,'' in \emph{Proceedings of the International
  Conference on Agricultural Engineering, Aarhus, Denmark}, 2016, pp. 26--29.

\bibitem{Grinblat2016vein}
G.~L. Grinblat, L.~C. Uzal, M.~G. Larese, and P.~M. Granitto, ``{Deep learning
  for plant identification using vein morphological patterns},''
  \emph{Computers and Electronics in Agriculture}, vol. 127, pp. 418--424,
  2016.

\bibitem{Dias2018}
P.~A. Dias, A.~Tabb, and H.~Medeiros, ``Apple flower detection using deep
  convolutional networks,'' \emph{Computers in Industry}, vol.~99, pp. 17--28,
  Aug. 2018.

\bibitem{Chen2016}
L.~C. Chen, G.~Papandreou, I.~Kokkinos, K.~Murphy, and A.~L. Yuille,
  ``{DeepLab: Semantic Image Segmentation with Deep Convolutional Nets, Atrous
  Convolution, and Fully Connected CRFs},'' \emph{IEEE Transactions on Pattern
  Analysis and Machine Intelligence}, vol.~PP, no.~99, pp. 1--1, 2018.

\bibitem{Tabb2018Datasets}
\BIBentryALTinterwordspacing
P.~A. Dias, A.~Tabb, and H.~Medeiros, ``\BIBforeignlanguage{en}{Data from:
  {Multi}-species fruit flower detection using a refined semantic segmentation
  network},'' 2018. [Online]. Available:
  \url{http://dx.doi.org/10.15482/USDA.ADC/1423466}
\BIBentrySTDinterwordspacing

\bibitem{Aggelopoulou2011color}
A.~D. Aggelopoulou, D.~Bochtis, S.~Fountas, K.~C. Swain, T.~A. Gemtos, and
  G.~D. Nanos, ``{Yield prediction in apple orchards based on image
  processing},'' \emph{Precision Agriculture}, vol.~12, no.~3, pp. 448--456,
  2011.

\bibitem{Hocevar2014color}
M.~Ho{\v{c}}evar, B.~{\v{S}}irok, T.~Gode{\v{s}}a, and M.~Stopar, ``{Flowering
  estimation in apple orchards by image analysis},'' \emph{Precision
  Agriculture}, vol.~15, no.~4, pp. 466--478, 2014.

\bibitem{Thorp2011}
K.~R. Thorp and D.~A. Dierig, ``{Color image segmentation approach to monitor
  flowering in lesquerella},'' \emph{Industrial Crops and Products}, vol.~34,
  no.~1, pp. 1150--1159, 2011.

\bibitem{Horton2017PeachAir}
R.~Horton, E.~Cano, D.~Bulanon, and E.~Fallahi, ``{Peach Flower Monitoring
  Using Aerial Multispectral Imaging},'' \emph{2016 ASABE International
  Meeting}, 2016.

\bibitem{Zeiler2014clarifai}
M.~D. Zeiler and R.~Fergus, ``Visualizing and understanding convolutional
  networks,'' in \emph{{European Conference on Computer Vision}}.\hskip 1em
  plus 0.5em minus 0.4em\relax Springer, 2014, pp. 818--833.

\bibitem{Hung2013Fruit}
C.~Hung, J.~Nieto, Z.~Taylor, J.~Underwood, and S.~Sukkarieh, ``{Orchard fruit
  segmentation using multi-spectral feature learning},'' \emph{IEEE
  International Conference on Intelligent Robots and Systems}, pp. 5314--5320,
  2013.

\bibitem{Bargoti2016Fruit}
S.~Bargoti and J.~P. Underwood, ``{Image Segmentation for Fruit Detection and
  Yield Estimation in Apple Orchards},'' \emph{Journal of Field Robotics},
  vol.~34, no.~6, pp. 1039--1060, 2017.

\bibitem{Long2015fcn}
J.~Long, E.~Shelhamer, and T.~Darrell, ``{Fully convolutional networks for
  semantic segmentation},'' in \emph{Proceedings of the IEEE Conference on
  Computer Vision and Pattern Recognition}, vol. 07-12-June, 2015, pp.
  3431--3440.

\bibitem{Garcia2017rev}
A.~Garcia-Garcia, S.~Orts-Escolano, S.~Oprea, V.~Villena-Martinez,
  P.~Martinez-Gonzalez, and J.~Garcia-Rodriguez, ``{A survey on deep learning
  techniques for image and video semantic segmentation},'' \emph{Applied Soft
  Computing}, vol.~70, pp. 41--65, 2018.

\bibitem{Krizhevsky2012alexnet}
A.~Krizhevsky, I.~Sutskever, and G.~E. Hinton, ``{ImageNet Classification with
  Deep Convolutional Neural Networks},'' \emph{Advances In Neural Information
  Processing Systems}, pp. 1--9, 2012.

\bibitem{Simonyan2014vgg}
K.~Simonyan and A.~Zisserman, ``Very deep convolutional networks for
  large-scale image recognition,'' \emph{CoRR (Presented at International
  Conference on Learning Representations, 2015)}, vol. abs/1409.1556, 2014.

\bibitem{Noh2015deconv}
H.~Noh, S.~Hong, and B.~Han, ``{Learning deconvolution network for semantic
  segmentation},'' in \emph{Proceedings of the IEEE International Conference on
  Computer Vision}, vol. 2015 Inter, 2015, pp. 1520--1528.

\bibitem{Kendall2015segnet}
V.~Badrinarayanan, A.~Kendall, and R.~Cipolla, ``Segnet: A deep convolutional
  encoder-decoder architecture for scene segmentation,'' \emph{IEEE
  Transactions on Pattern Analysis and Machine Intelligence}, 2017.

\bibitem{Hariharan2015hyper}
B.~Hariharan, P.~Arbel{\'{a}}ez, R.~Girshick, and J.~Malik, ``{Hypercolumns for
  object segmentation and fine-grained localization},'' in \emph{Proceedings of
  the IEEE Conference on Computer Vision and Pattern Recognition}, vol.
  07-12-June, 2015, pp. 447--456.

\bibitem{He2016resnet}
K.~He, X.~Zhang, S.~Ren, and J.~Sun, ``Deep residual learning for image
  recognition,'' in \emph{Proceedings of the IEEE Conference on Computer Vision
  and Pattern Recognition}, 2016, pp. 770--778.

\bibitem{Everingham2015pascal}
M.~Everingham, S.~A. Eslami, L.~Van~Gool, C.~K. Williams, J.~Winn, and
  A.~Zisserman, ``{The PASCAL Visual Object Classes challenge: A
  retrospective},'' \emph{International journal of computer vision}, vol. 111,
  no.~1, pp. 98--136, 2015.

\bibitem{Lin2014coco}
T.-Y. Lin, M.~Maire, S.~Belongie, J.~Hays, P.~Perona, D.~Ramanan,
  P.~Doll{\'a}r, and C.~L. Zitnick, ``{Microsoft COCO: Common objects in
  context},'' in \emph{{European Conference on Computer Vision}}.\hskip 1em
  plus 0.5em minus 0.4em\relax Springer, 2014, pp. 740--755.

\bibitem{Krahenbuhl2012}
P.~Kr{\"{a}}henb{\"{u}}hl and V.~Koltun, ``{Efficient Inference in Fully
  Connected CRFs with Gaussian Edge Potentials},'' in \emph{Advances in neural
  information processing systems}, 2011, pp. 109--117.

\bibitem{Dias2018rgr}
\BIBentryALTinterwordspacing
P.~A. Dias and H.~Medeiros, ``{Semantic Segmentation Refinement by Monte Carlo
  Region Growing of High Confidence Detections},'' in \emph{ArXiv}, 2018.
  [Online]. Available: \url{https://arxiv.org/abs/1802.07789}
\BIBentrySTDinterwordspacing

\bibitem{Caesar2017stuff}
H.~Caesar, J.~Uijlings, and V.~Ferrari, ``{COCO-Stuff: Thing and Stuff Classes
  in Context},'' in \emph{Proceedings of the IEEE Conference on Computer Vision
  and Pattern Recognition}, 2018.

\bibitem{Girshick2014rcnn}
R.~Girshick, J.~Donahue, T.~Darrell, and J.~Malik, ``Rich feature hierarchies
  for accurate object detection and semantic segmentation,'' in
  \emph{Proceedings of the IEEE conference on Computer Vision and Pattern
  Recognition}, 2014, pp. 580--587.

\bibitem{Jia2014a}
Y.~Jia, E.~Shelhamer, J.~Donahue, S.~Karayev, J.~Long, R.~Girshick,
  S.~Guadarrama, and T.~Darrell, ``Caffe: Convolutional architecture for fast
  feature embedding,'' in \emph{Proceedings of the 22nd ACM International
  Conference on Multimedia}.\hskip 1em plus 0.5em minus 0.4em\relax New York,
  NY, USA: ACM, 2014, pp. 675--678.

\end{thebibliography}

\end{document}